\documentclass[journal]{IEEEtran}

\usepackage{cite}
\usepackage{xcolor}
\usepackage{amsmath}
\usepackage{graphicx}
\usepackage{textcomp}
\usepackage{algorithmic}
\usepackage[hidelinks]{hyperref}
\usepackage{amsmath,amssymb,amsfonts}

\hypersetup{colorlinks=true,citecolor=blue,
            linkcolor=blue,urlcolor=blue}
\graphicspath{{.}}

\def\BibTeX{{\rm B\kern-.05em{\sc i\kern-.025em b}\kern-.08em
    T\kern-.1667em\lower.7ex\hbox{E}\kern-.125emX}}
\begin{document}

\title{Models Developed for Spiking Neural Networks}

\author{
\IEEEauthorblockN{
Shahriar Rezghi Shirsavar\IEEEauthorrefmark{2}\IEEEauthorrefmark{3},
Abdol-Hossein Vahabie\IEEEauthorrefmark{2},
Mohammad-Reza A. Dehaqani\IEEEauthorrefmark{2}\IEEEauthorrefmark{3}\IEEEauthorrefmark{1}
}

\IEEEauthorblockA{\IEEEauthorrefmark{2}School of Electrical and Computer Engineering, College of Engineering, University of Tehran, Tehran, Iran\\\{shahriar.rezghi, h.vahabie, dehaqani\}@ut.ac.ir}

\IEEEauthorblockA{\IEEEauthorrefmark{3}School of Cognitive Sciences, Institute for Research in Fundamental Sciences (IPM), Tehran, Iran}

\IEEEauthorblockA{\IEEEauthorrefmark{1}Corresponding author: Mohammad-Reza A. Dehaqani, dehaqani@ut.ac.ir}
}

\maketitle

\begin{abstract} \label{sec:abstract}
Emergence of deep neural networks (DNNs) has raised enormous attention towards artificial neural networks (ANNs) once again. They have become the state-of-the-art models and have won different machine learning challenges. Although these networks are inspired by the brain, they lack biological plausibility, and they have structural differences compared to the brain. Spiking neural networks (SNNs) have been around for a long time, and they have been investigated to understand the dynamics of the brain. However, their application in real-world and complicated machine learning tasks were limited. Recently, they have shown great potential in solving such tasks. Due to their energy efficiency and temporal dynamics there are many promises in their future development. In this work, we reviewed the structures and performances of SNNs on image classification tasks. The comparisons illustrate that these networks show great capabilities for more complicated problems. Furthermore, the simple learning rules developed for SNNs, such as STDP and R-STDP, can be a potential alternative to replace the backpropagation algorithm used in DNNs.
\end{abstract}


\section{Introduction} \label{sec:introduction}

Artificial neural networks (ANNs) have been around for a long time. The first generation of ANNs was made of McCulloch-Pitts \cite{mcculloch_logical_1943} neurons. These neurons were the result of simple modeling of the biological neurons. These neurons work with binary signals by accumulating the input and firing a spike when their internal potential reaches a certain threshold. Despite their simplicity, these neurons are powerful and are used to create multi-layer perceptron networks. The second generation of ANNs used neuron models with continuous activation functions (sigmoid \cite{rumelhart_learning_1986} and ReLU \cite{nair_rectified_2010}, for example). Feedforward and recurrent networks are a part of this generation. These networks work well with analog signals and are able to approximate analog functions quite well.

In recent years, the second generation of ANNs has had major success and has been able to outperform other methods in most areas \cite{collobert_natural_2011, graves_framewise_2005, krizhevsky_imagenet_2017}. Although these networks are able to achieve high accuracies, they are data- \cite{sun_revisiting_2017} and energy-hungry \cite{li_evaluating_2016} and have lower resistance to noise and disturbance \cite{alcorn_strike_2019}. The human brain is a very capable neural network. It is able to learn with a few samples and has high generalizability. It is able to store a large amount of information and has amazing energy efficiency. Mentioned problems of DNNs have led researchers to come up with a biologically plausible structure that models the brain more closely compared to the second these networks. Spiking neural networks (SNNs) are the third generation of ANNs and have a higher biological plausibility compared to DNNs. These networks use spatio-temporal information, much like biological neurons and propagate binary spike trains instead of analog signals. Several electrophysiological studies emphasize the role of temporal dynamics in neural coding \cite{dehaqani_temporal_2016, dehaqani_selective_2018}. The binary signals passed through these networks can have a high sparsity rate that is directly related to energy efficiency. There have also been theoretical discussions about the noise invariance of SNNs \cite{beer_why_2020}.

The goal of this review is to introduce different models developed for spiking neural networks and to compare them. First, we will introduce the building blocks of SNNs in Section \ref{sec:building-blocks} and explain their components. Then, successful networks will be introduced in Section \ref{sec:snn-models} and discussed. Finally, the review will be concluded in Section \ref{sec:conclusion}, and possible future directions will be explored.

\section{Building Blocks} \label{sec:building-blocks}

\begin{figure*}[htbp]
\centerline{\includegraphics[width=\textwidth]{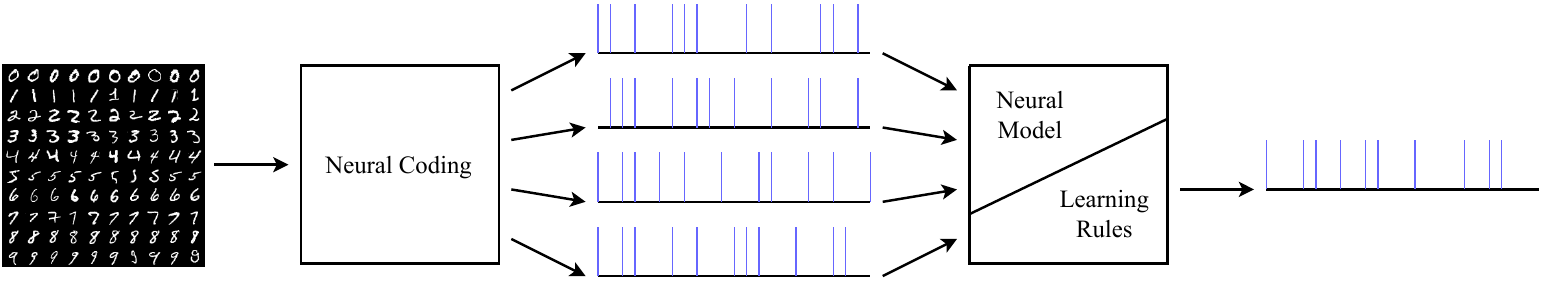}}
\caption{A schematic for a spiking neural network process. The input signals are converted to spike trains using a coding scheme and then forwarded through a network that is made of models of the biological neurons. Learning happens in the network, and the network outputs are used for various tasks.}
\label{fig:snn-overview}
\end{figure*}

SNNs have different building blocks. Each of these blocks model a specific functionality of the brain. Since different models have been proposed for each building block, a singly defined structure does not exist for SNNs. Instead, the choice of which component to use for a building block and the connection between different components is important to have a successful SNN model. These building blocks are explored in Sections \ref{sec:block-neural}, \ref{sec:block-coding}, and \ref{sec:block-learning}. A demonstration of an SNN can be seen in Figure \ref{fig:snn-overview}.

\subsection{Models of Biological Neurons} \label{sec:block-neural}

Artificial neural networks are made of modeled neurons. They model biological neurons, and in order to have a biologically plausible network, the model has to follow the spatio-temporal dynamics of a real neuron. However, biological neuron has complex dynamics, and complicated models are computationally demanding. Since the simulated network can be made of millions of neurons, using simpler models can lead to an efficient simulation. Several models have been proposed that vary in their level of complexity and computational weight. Some of these models will be introduced in Sections \ref{sec:neural-lif}, \ref{sec:neural-srm}, and \ref{sec:neural-izh}.

\subsubsection{Leaky Integrate-and-fire} \label{sec:neural-lif}

Leaky integrate-and-fire (LIF) \cite{abbott_lapicques_1999} is a widely used model. If we want to model the neuron in a simple manner, the inputs are aggregated, and once the potential passes a threshold, the neuron fires. It is assumed that the shape of the spikes is generally the same in the LIF model, and the shape of the spike does not contain information. Instead, the information is transmitted in the presence and absence of the spikes. This model does not try to reconstruct the shape of the spike and models the neuron with a combination of a capacitator and a resistor. When the internal potential passes a certain threshold, a spike is fired, and the potential of the capacitator reverts to the resting state. The differential equation of this model is as follows:
\begin{equation} \label{eq:lif-equation}
C \frac{dV}{dt} = -g_L (V(t) - E_L) + I(t)
\end{equation}
where $C$ is the capacitance and $g_L$ is the resistance.

\subsubsection{Spike-response Model} \label{sec:neural-srm}

Differential equations are often used in the modeling of biological neurons. However, in the spike-response model (SRM) \cite{jolivet_spike_2003}, the parameters of the model are replaced with functions of time. This model is a generalized version of the LIF model. It can have both fixed and variable thresholds compared to the LIF model, which only has a fixed threshold. The modeling of the subthreshold activity of the SRM model is better compared to the LIF model. However, it has a simpler modeling of the spike compared to more complex models. The equation of this model is shown below:
\begin{equation} \label{eq:srm-equation}
u(t) = \eta(t-\hat{t}) + \int_0^\infty \kappa(t-\hat{t}, s)I(t-s)ds
\end{equation}

In this equation, $\hat{t}$ is the time of the previous spike, $\eta$ function determines the shape of the action potential, and $\kappa$ function determines the linear response to the delta function as input.

\subsubsection{Izhikevich Model} \label{sec:neural-izh}

The Izhikevich model reproduces the spiking pattern of cortical neurons \cite{izhikevich_simple_2003}. This model combines the biological plausibility of Hodgkin-Huxley models \cite{hodgkin_quantitative_1952} (a complicated model of spiking neurons) and the computational efficiency of integrate-and-fire neurons \cite{izhikevich_simple_2003}. Having such features makes this type of model suitable for SNNs. This model can be formulated as follows:
\begin{align}
v^\prime &= 0.04v^2 + 5v + 140 - u + I \\
u^\prime &= a(bv - u)
\end{align}
with the auxiliary after-spike resetting
\begin{equation}
\text{if } v \ge 30mV\text{, then}
\begin{cases}
v &\leftarrow c \\
u &\leftarrow u + d
\end{cases}
\end{equation}
where $a$, $b$, $c$, and $d$ are dimensionless parameters, $v$ is the membrane potential, and $u$ is the membrane recovery variable, which accounts for the behavior of ion channels and provides negative feedback to $v$ \cite{izhikevich_simple_2003}.

\subsection{Neural Coding} \label{sec:block-coding}

Input signals to spiking neural networks can be analog, much similar to the signals captured by human sensors (audio and images, for example) \cite{kostakos_human_2017}. However, the modeled neurons use spikes as both input and output. In order for these networks to be able to process the analog input signals, we need to apply conversion schemes. Some of these coding schemes will be explained in Sections \ref{sec:coding-rate}, \ref{sec:coding-temporal}, and \ref{sec:coding-phase}.

\subsubsection{Rate Coding} \label{sec:coding-rate}

Rate coding \cite{adrian_impulses_1926} is a frequently used coding mechanism, and the information is transmitted through the rate of firing of the neurons. This mechanism uses the intensity of the input signal (for example, in an image input, the value of a pixel is its intensity) for conversion, and the rate of firing is correlated to the intensity of the input. The Poisson processes can be used to model this coding scheme \cite{wiener_decoding_2003}, and the mean of the distribution is the rate of firing.

\subsubsection{Temporal Coding} \label{sec:coding-temporal}

A lengthy window of time is needed to transfer information using rate coding, and the signal has low sparsity. Another suggested method of coding is temporal coding \cite{buzsaki_rhythms_2006}. In this coding, the information is transmitted using the timing of the spikes. One form of this coding scheme is that the inputs that have higher values are translated into earlier firing times (see \cite{kheradpisheh_stdp-based_2018, mozafari_bio-inspired_2019} for implementations in SNN). This coding is fast and sparse, and this sparsity can speed up SNN simulations. This coding has multiple forms, such as time-to-first-spike \cite{johansson_first_2004}, rank order coding \cite{thorpe_rank_1998}, and relative spike latency \cite{gollisch_rapid_2008}. It has been suggested \cite{gautrais_rate_1998} that temporal coding might be more efficient in some situations. In rank order coding, the exact timing of the spikes is not considered, and the timings are only considered relative to each other.

\subsubsection{Phase Coding} \label{sec:coding-phase}

Phase coding \cite{okeefe_phase_1993, laurent_dynamical_1996} is another coding scheme that encodes information in spike patterns that have phases correlating with internally generated background oscillation rhythms. In this coding, the neurons fire in different phases, and the phase is used to transmit information.

\subsection{Learning} \label{sec:block-learning}

Learning strategies are an important part of a neural network. Certain learnable parameters of the network get tuned with the goal of learning robust and generalizable representations. Different learning rules for SNNs will be explored in Sections \ref{sec:learning-stdp}, \ref{sec:learning-rstdp}, \ref{sec:learning-backprop}, and \ref{sec:learning-conversion}.

\subsubsection{STDP} \label{sec:learning-stdp}

Although deep neural networks can achieve high accuracies with backpropagation \cite{tan_efficientnet_2020, collobert_natural_2011, graves_framewise_2005, krizhevsky_imagenet_2017}, their convergence is slow, and they need a large number of labeled examples and energy in order to learn. But humans are able to learn with a few examples and without explicit labels and a small amount of energy. In the brain, learning occurs through changes in the strength of the connection between neurons. This change in the strength of connections is called synaptic plasticity \cite{citri_synaptic_2008}.  Different learning rules have been developed that take advantage of synaptic plasticity. One widely recognized method is the spike-timing-dependent plasticity (STDP) learning rule \cite{markram_regulation_1997}. This method works by adjusting synaptic weights using the timing of the spikes. In this rule, when a pre-synaptic neuron fires before a post-synaptic neuron, their connection is strengthened, and when a pre-synaptic neuron fires after a post-synaptic neuron, their connection is weakened. Changing these synaptic weights can change the spike flow of a network \cite{vreeken_spiking_2003}. STDP allows the neurons to extract and learn features that are frequently seen in the input. STDP is an asymmetrical form of the Hebbian learning rule \cite{hebb_organization_1949}. The Hebbian learning rule suggests that if a pre-synaptic neuron repeatedly or persistently takes part in the firing of the post-synaptic neuron, then their connection should be strengthened \cite{hebb_organization_1949}.  STDP takes a step further and adds a mechanism to weaken the connection, as explained earlier.

STDP is an unsupervised learning rule \cite{markram_regulation_1997}. It can be formulated this way:
\begin{equation}
\Delta W_{i,j} =
\begin{cases}
A^{+} exp(-\frac{\Delta t}{\tau^+}), & t_{pre} < t_{post} \\
A^{-} exp(+\frac{\Delta t}{\tau^-}), & t_{pre} \geq t_{post} \\
\end{cases}
\end{equation}
Where $A^{+}$, $A^{-}$ are the positive learning rate, negative learning rate and $\Delta t = t_{post} - t_{pre}$, respectively.

\subsubsection{R-STDP} \label{sec:learning-rstdp}

There is also a reinforcement learning rule that uses STDP, and it is called reward-modulated STDP (R-STDP) \cite{fremaux_neuromodulated_2016}. Reinforcement learning helps machine learning models to learn and make a series of decisions in an environment. In this rule, the agent learns to achieve a defined goal in an uncertain and complex environment. The R-STDP learning rule changes the STDP method such that the neurons that have the correct response are rewarded, and the neurons that have the incorrect response are punished. Research suggests that when the input signals have frequent features that are not helpful in the decision-making process, the R-STDP rule can learn to ignore these features and improve the decision-making process \cite{mozafari_bio-inspired_2019}.

One way of formulating the R-STDP learning rule is as follows:
\begin{equation}
\Delta W_{i,j} =
\begin{cases}
\begin{cases}
A^{+}_r exp(-\frac{\Delta t}{\tau^+}_r), & t_{pre} < t_{post} \\
A^{-}_r exp(+\frac{\Delta t}{\tau^-}_r), & t_{pre} \geq t_{post} \\
\end{cases}, & \text{if rewarded} \\ \\
\begin{cases}
A^{-}_p exp(-\frac{\Delta t}{\tau^-}_p), & t_{pre} < t_{post} \\
A^{+}_p exp(+\frac{\Delta t}{\tau^+}_p), & t_{pre} \geq t_{post} \\
\end{cases}, & \text{if punished} \\
\end{cases}
\end{equation}
Where $A^{+}_r$, $A^{-}_r$, $A^{-}_p$, $A^{+}_p$ are reward positive, reward negative, punishment negative and punishment positive learning rates and $\Delta t = t_{post} - t_{pre}$, respectively.

\subsubsection{Backpropagation} \label{sec:learning-backprop}

Backpropagation is a learning rule that is widely used today in deep neural networks. However, there are some problems when one tries to apply this rule to spiking neural networks. Since spike trains are not differentiable, using backpropagation is not straightforward. Despite this difficulty, various supervised learning rules have been developed for SNNs that use backpropagation (see Section \ref{sec:snn-models}). SpikeProp \cite{bohte_spikeprop_2000} can be thought of as the first backpropagation algorithm developed for SNNs. This method works by propagating the temporal error at the output. It is able to overcome the discontinuity associated with thresholding by linearizing the relationship between the post-synaptic input and the output spike time. Backpropagation methods are generally thought of as having less biological plausibility.

\subsubsection{ANN-to-SNN} \label{sec:learning-conversion}

The combination of gradient descent and error backpropagation is the primary learning rule in deep neural networks. This method has been quite successful. As mentioned, these methods have been used in spiking neural networks after some modifications. However, there is another way to use this method to have a trained network. One can train a deep network, convert it to an SNN, and then go on to perform inference operations using the SNN. It is critical to consider which layers to replace in the deep network and what changes to make (see Section \ref{sec:model-conversion}) to not use accuracy in the resulting network (or have a very small drop of accuracy) \cite{diehl_fast-classifying_2015}.

\subsection{Libraries and Toolboxes} \label{block-toolbox}

A number of high-performance and established libraries and tools have been developed for DNNs. Some examples of these tools are PyTorch \cite{paszke_pytorch_2019}, TensorFlow \cite{developers_tensorflow_2022}, and MXNet \cite{chen_mxnet_2015}. These tools have helped researchers work faster and reach better results. Some libraries are also developed for SNNs, but these libraries are not comparable to DNN simulation tools.  Some of the simulation tools for SNNs are written on top of PyTorch. Examples of these libraries are BindsNET \cite{hazan_bindsnet_2018}, SpykeTorch \cite{mozafari_spyketorch_2019}, and Norse \cite{pehle_norse_2021}. Others like Brian \cite{stimberg_brian_2019} and NEST \cite{gewaltig_nest_2007} are written from scratch. One notable library written from the ground up is the Spyker \footnote{\url{https://github.com/ShahriarRezghi/Spyker}} library. It uses highly optimized low-level backends on both CPU and GPU devices, has both C++ and Python interfaces, is multiple times faster compared to its predecessors, and has the ability to be integrated with commonly used tools like PyTorch and Numpy. This library supports rank order coding, rate coding, integrate-and-fire neurons, STDP, R-STDP, and backpropagation (experimental).

\begin{table*}[htbp]
\caption{Considered papers and some of their characteristics including accuracy on the MNIST dataset, learning method, coding scheme, and whether convolution is used or not.}
\label{tab:selected-papers}

\begin{center}
\begin{tabular}{|c|c|c|c|c|}
\hline
Paper                               & Average Accuracy (\%) & Method               & Coding Scheme  & Convolutional \\ \hline
\cite{shirsavar_faster_2022}        & 99.42 (99.01)         & STDP+SVM             & Temporal       & Yes           \\ \hline
\cite{kheradpisheh_stdp-based_2018} & 98.40                 & STDP+SVM             & Temporal       & Yes           \\ \hline
\cite{drix_sparse_2020}             & 98.10                 & Somato-dendritic+SVM & Rate           & No            \\ \hline
\cite{mozafari_bio-inspired_2019}   & 97.20                 & STDP+R-STDP          & Temporal       & Yes           \\ \hline
\cite{xu_improving_2017}            & 95.20                 & R-STDP               & -              & No            \\ \hline
\cite{saunders_locally_2019}        & 95.07                 & STDP+Voting          & Rate           & Yes           \\ \hline
\cite{beyeler_categorization_2013}  & 91.64                 & STDP-Like            & Rate           & No            \\ \hline
\cite{xu_deep_2020}                 & 91.40                 & STDP                 & Rate           & Yes           \\ \hline
\cite{wang_compsnn_2021}            & 91.22                 & Tempotron+Voting     & Multi-temporal & No            \\ \hline
\cite{ding_optimal_2021}            & 99.46                 & ANN-to-SNN           & Rate           & Yes           \\ \hline
\cite{rueckauer_conversion_2017}    & 99.44                 & ANN-to-SNN           & Rate           & Yes           \\ \hline
\cite{diehl_fast-classifying_2015}  & 99.14 (98.68)         & ANN-to-SNN           & Rate           & Yes           \\ \hline
\cite{zhang_fast_2019}              & 98.40                 & ANN-to-SNN           & Rate           & Both          \\ \hline
\cite{shen_backpropagation_2022}    & 99.67                 & Backpropagation      & Rate           & Yes           \\ \hline
\cite{lee_enabling_2020}            & 99.59                 & Backpropagation      & Rate           & Yes           \\ \hline
\cite{zhao_backeisnn_2021}          & 99.58                 & Backpropagation      & Rate           & Yes           \\ \hline
\cite{zhang_spike-train_2019}       & 99.57                 & Backpropagation      & Rate           & Yes           \\ \hline
\cite{zhang_temporal_2020}          & 99.50                 & Backpropagation      & Rate           & Yes           \\ \hline
\cite{wu_spatio-temporal_2018}      & 99.42 (98.89)         & Backpropagation      & Rate           & Both          \\ \hline
\cite{cheng_lisnn_2020}             & 99.42                 & Backpropagation      & Rate           & Yes           \\ \hline
\cite{lee_training_2016}            & 99.31                 & Backpropagation      & Rate           & Yes           \\ \hline
\cite{kheradpisheh_temporal_2020}   & 97.40                 & Backpropagation      & Temporal       & No            \\ \hline
\cite{kheradpisheh_bs4nn_2022}      & 97.00                 & Backpropagation      & Temporal       & No            \\ \hline
\end{tabular}
\end{center}
\end{table*}

\section{Developed Models} \label{sec:snn-models}

The initial set of papers considered for this review with descent accuracy and structure are listed in Table \ref{tab:selected-papers}. These papers are mainly gathered from ScienceDirect and PubMed databases. Since the accuracies of different models will be compared, the search needs to include papers that solve the same dataset. Since the MNIST dataset is a widely used benchmark dataset, it can be used to compare the performance of different models. The review aims to find the best-performing models developed for SNNs and the models that perform better than others that have similar building blocks will be explored further in Sections \ref{sec:model-stdp}, \ref{sec:model-rstdp}, \ref{sec:model-backprop}, and \ref{sec:model-conversion}. Furthermore, a detailed summary of the structure and the features of the selected papers is shown in Table \ref{tab:papers-summary}.

\subsection{STDP Network} \label{sec:model-stdp}

\begin{figure*}[htbp]
\centerline{\includegraphics[width=\textwidth]{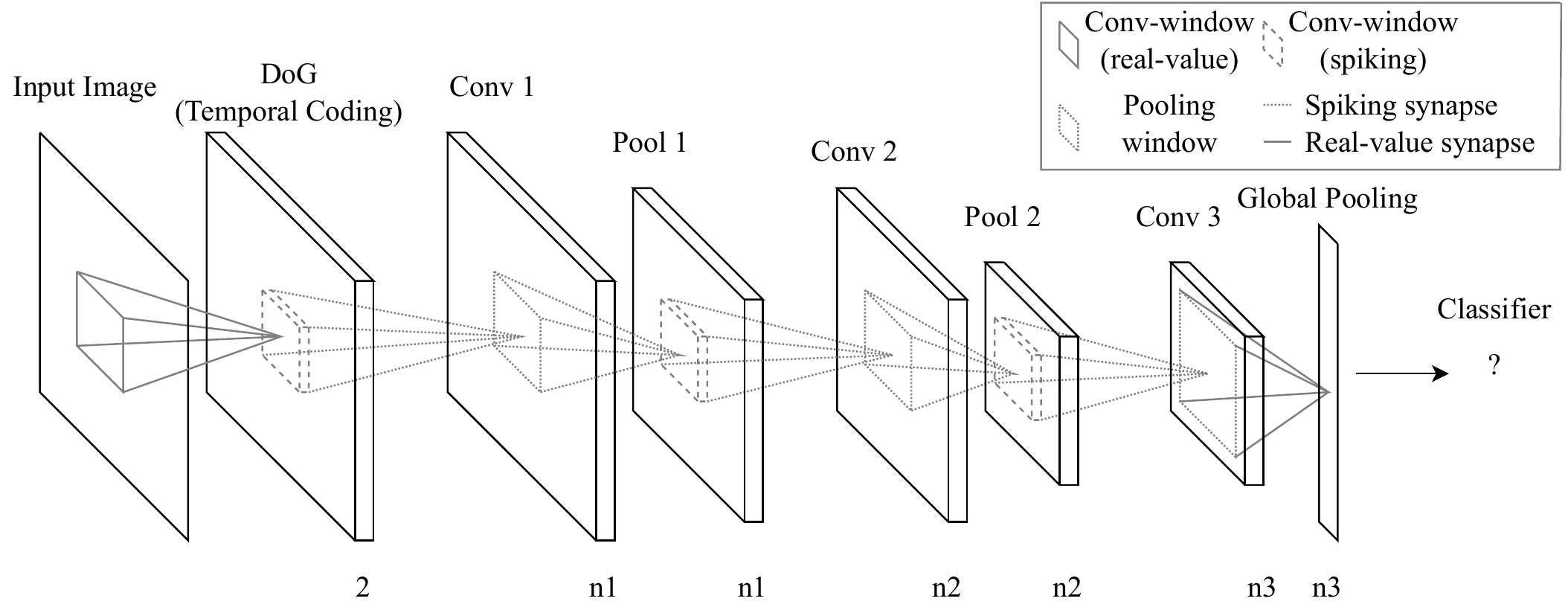}}
\caption{Schematics for the Kheradpisheh et al. network. This network uses an on-center and an off-center DoG filters on the inputs with rank order coding afterwards, has two convolutional layers, each followed by a pooling layer, and an SVM classifier at the end. Please note that the third convolutional layer is not used on the MNIST dataset.}
\label{fig:kherad-network}
\end{figure*}

The Kheradpisheh et al. network \cite{kheradpisheh_stdp-based_2018} combines the STDP learning rule with an external classifier. The structure for the MNIST dataset is made of two convolutional layers followed by an SVM classifier (see schematics in Figure \ref{fig:kherad-network}). The input is filtered with an on- and an off-center DoG filter, and temporal coding is used. Each convolutional layer is followed by an IF activation function and a pooling layer. The layers are trained with the STDP learning rule, and lateral inhibition and WTA mechanisms are employed. The training of the layers is done in sequence, and after the network is trained, the network outputs for the training and the testing set are computed, and an SVM classifier is used to classify the outputs. This network reaches an impressive testing accuracy of 98.40\%. This structure works well but has less biological plausibility compared to the Mozafari et al. network \cite{mozafari_bio-inspired_2019} (explained in Section \ref{sec:model-rstdp}) due to the usage of the SVM classifier, which has no biological roots.

\begin{figure*}[htbp]
\centerline{\includegraphics[width=\textwidth]{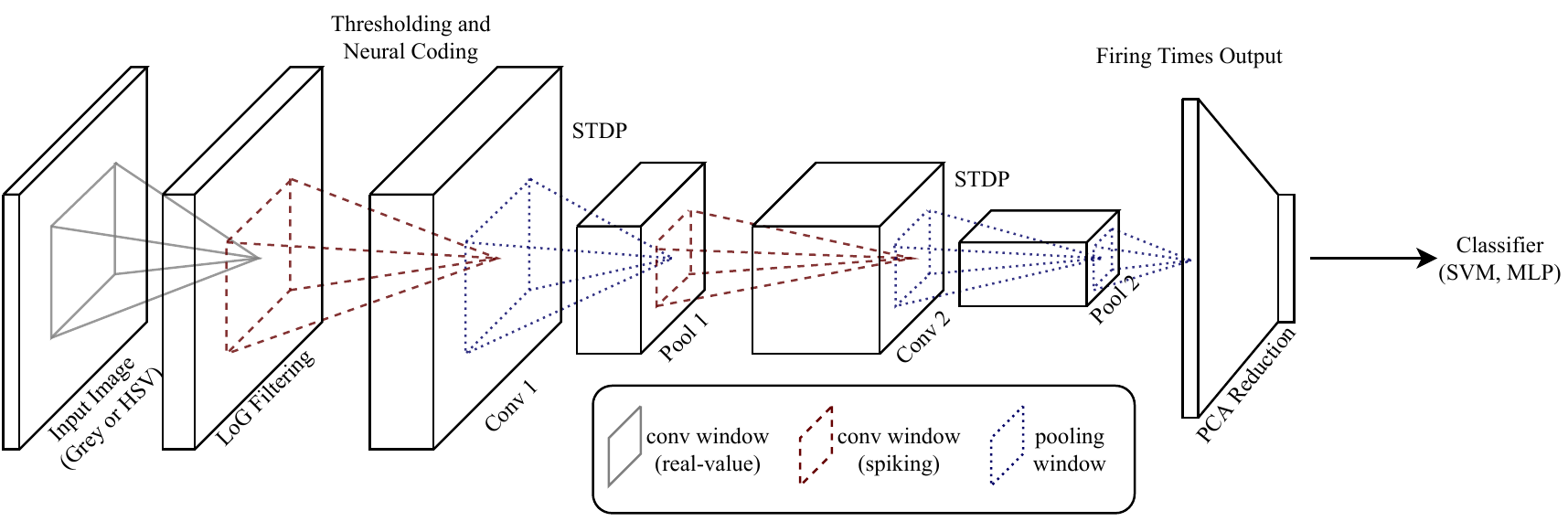}}
\caption{Overview of Shirsavar et al. Network. This network uses three LoG filters with rank order coding. It has two convolutional layers, each followed by a pooling layer. Finally, PCA dimension reduction is used on the network outputs,and the SVM classifier is run.}
\label{fig:shirsavar-network}
\end{figure*}

The work of Kheradpisheh et al. can be further improved, and the aim of the Shirsavar et al. network \cite{shirsavar_faster_2022} is to do so. This work increases the accuracy of previous works while improving the speed of processing by orders of magnitude (a visualization of this network can be seen in Figure \ref{fig:shirsavar-network}). It is based on previous networks and makes some changes. The first change happens in the output of the network. The outputs are changed from binary indication of spikes to the firing time of the output neurons. The following change is to introduce a dimension reduction before the SVM classification process. This dimension reduction (PCA) has little to no effect on the accuracy, while it improves the runtime significantly. The final change consists of an improved training process. With better control of learning parameters and quantizing the weights afterward, the training iterations can be significantly reduced (only one iteration for the MNIST dataset). This network reaches 99.011$\pm$0.068 percent accuracy with a linear classifier and 99.421$\pm$0.033 percent with a non-linear classifier. The biological plausibility of this network is much like the Kheradpisheh et al. Network.

\subsection{R-STDP Network} \label{sec:model-rstdp}

\begin{figure*}[htbp]
\centerline{\includegraphics[width=\textwidth]{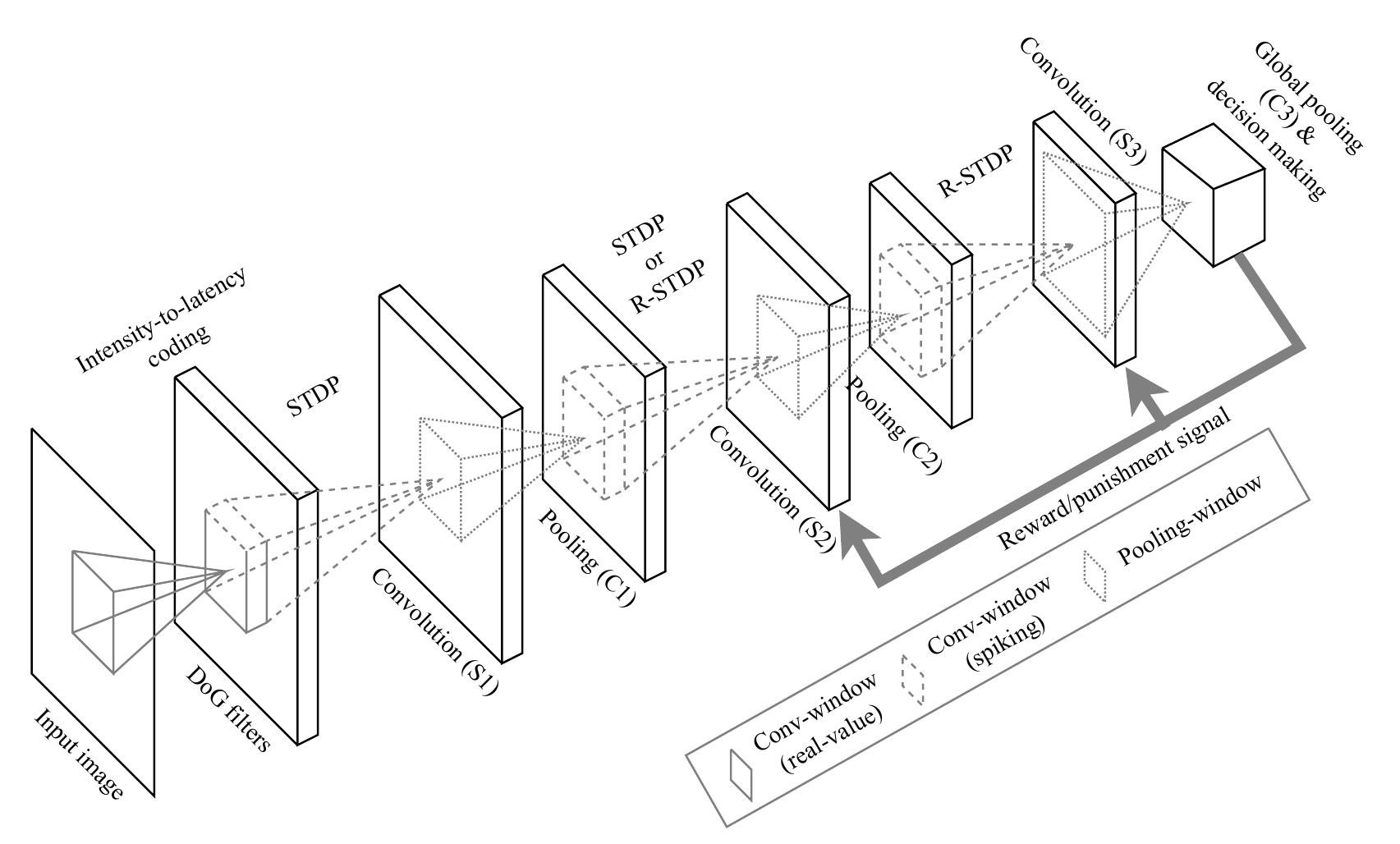}}
\caption{Schematics for the Mozafari et al. network. The network uses three on-center and three off-center DoG filters on inputs with rank order coding afterward, has three convolutional layers, each followed by a pooling layer, and the STDP and R-STDP learning rules are used to train the network. The network has a native classifier.}
\label{fig:mozaf-network}
\end{figure*}

STDP is an unsupervised learning rule, and it alone is not enough to solve classification tasks. The Mozafari et al. \cite{mozafari_bio-inspired_2019} network solves this problem by combining it with the R-STDP learning rule. It proposes a network structure that has several components (see schematics in Figure \ref{fig:mozaf-network}). The structure includes three on- and off-center Difference of Gaussian (DoG) filters on the inputs for feature enhancement. The network uses a temporal coding scheme and has three convolutional layers, each with an integrate-and-fire (IF) activation function and a pooling operation. The training of the layers is done in sequence (one layer after another). Lateral inhibition is used in this network and is an essential part of the learning process. In this mechanism, when a neuron fires in a feature map, it stops other neurons in the same position belonging to other feature maps from firing. After inhibition, a number of neurons are selected using a winner-take-all (WTA) mechanism, and the STDP learning rule is applied to them. Lateral inhibition and WTA force the neurons to compete with each other in order to learn. The first and the second layers use the STDP learning rule. However, the third layer uses the R-STDP learning rule, and its activation function has an infinite threshold. The infinite threshold stops the neurons from firing, and the internal potentials of the neurons are used instead. In this layer, output channels are assigned to different output classes, and the neuron with the highest potential is selected. If the mapped class label of the channel that the neuron belongs to is the same as the class label of the input image, then the neuron will be rewarded (see Section \ref{sec:learning-rstdp}). If the labels are not the same, it will be punished. This structure has a high biological plausibility due to the fact that it uses DoG filters, STDP and R-STDP learning rules and does not use a classifier with no biological roots (SVM, for example). It achieves a good test accuracy of 97.20\%.

\begin{table*}[htbp]
\caption{In this table, explained papers and their highlighted features are summarized and their accuracies, biological plausibility, and sparsity levels are mentioned.}
\label{tab:papers-summary}

\begin{center}
\begin{tabular}{|c|c|c|c|c|}
\hline
Paper                               & Description                                                                                                                                                                                                                                                                                                            & Average Accuracy(\%) & Biological Plausibility & Sparsity \\ \hline
\cite{mozafari_bio-inspired_2019}   & \begin{tabular}[c]{@{}c@{}}DoG feature enhancement\\ Temporal coding\\ Three convolution layers with IF, lateral inhibition\\ STDP in first two layers, R-STDP in third\\ Classification using assigning channels to labels\end{tabular}                                                                               & 97.20                & High                    & High     \\ \hline
\cite{kheradpisheh_stdp-based_2018} & \begin{tabular}[c]{@{}c@{}}DoG feature enhancement\\ Temporal coding\\ Two convolution layers with IF, lateral inhibition\\ STDP learning rule\\ Classification using SVM\end{tabular}                                                                                                                                 & 98.40                & Medium                  & High     \\ \hline
\cite{shirsavar_faster_2022}        & \begin{tabular}[c]{@{}c@{}}Based on previous networks\\ Neuron firing times as output\\ PCA before SVM to improve runtime speed\\ Network weight quantization\end{tabular}                                                                                                                                             & 99.42 (99.01)        & Medium                  & High     \\ \hline
\cite{diehl_fast-classifying_2015}  & \begin{tabular}[c]{@{}c@{}}Convert ANN to SNN\\ ReLU and dropout in ANN\\ Normalizing of weights to close accuracy gap\\ Two convolution Layers, one fully connected layer\end{tabular}                                                                                                                                & 99.14                & Low                     & Low      \\ \hline
\cite{ding_optimal_2021}            & \begin{tabular}[c]{@{}c@{}}Train weight normalization parameters\\ Replace ReLU with rate norm layer (RNL)\\ Add a parameter to the loss function\end{tabular}                                                                                                                                                         & 99.46                & Low                     & Low      \\ \hline
\cite{lee_training_2016}            & \begin{tabular}[c]{@{}c@{}}Train SNN with backpropagation\\ LIF with lateral inhibition\\ Treat spikes as continuous signals\\ Normalizing weights to stabilize gradients\\ Weight and threshold regularization\\ Adam optimizer\\ Data augmentation\\ two convolution layers, two fully connected layers\end{tabular} & 99.31                & Low                     & Low      \\ \hline
\cite{shen_backpropagation_2022}    & \begin{tabular}[c]{@{}c@{}}Use average firing rate to determine output label\\ Train with mean squared error\\ Only calculate the gradient at the moment of spike\\ Add residual connections between spikes in the backward path\end{tabular}                                                                          & 99.67                & Low                     & Low      \\ \hline
\end{tabular}
\end{center}
\end{table*}

\subsection{Backpropagation Network} \label{sec:model-backprop}

Backpropagation has been the primary learning rule in DNNs, and one approach that is researched is to apply backpropagation directly to an SNN network. Since spikes passed through the network are not differentiable because of discontinuous step jumps, applying backpropagation is not as straightforward, but Lee et al. \cite{lee_training_2016} propose a method to solve this problem. In this work, leaky integrate-and-fire neural model and lateral inhibition are used to model biological neurons and introduce competition in the network architecture, respectively. This method ignores fluctuations in the spikes and treats them as continuous signals which helps it derive the necessary error gradients for the backpropagation.  Weight normalization is also used to avoid vanishing or exploding gradients as the network becomes deeper (has more layers). Weights are uniformly initialized, and the threshold is calculated with respect to the number of synapses of each neuron. Weight decay and threshold regularization are used in this model to improve stability and generalizability. Adam optimization \cite{kingma_adam_2017} is employed here, and the network has two convolutional layers, each following a pooling operation. The network also uses two fully connected layers to compute the output of the model. This structure achieves a high testing accuracy of 99.31\%. However, data augmentation is used to improve the results, and the accuracy without the augmentation is not reported. Since backpropagation is used, the network lacks biological plausibility.

Another work that reaches state-of-the-art results is the Shen et al. network \cite{shen_backpropagation_2022}. This work uses the average firing rate of the last layer to determine the output label and trains the network with mean squared error. Previous works based on surrogate-gradient methods calculate the gradients around the threshold. This method does not account for the fact that the neurons not emiting spikes in the forward path will participate in the parameter update process. Also, earlier spike moments will have a larger influence on the weight update compared to later ones. This conflicts with the neurophysiology rules \cite{shen_backpropagation_2022}. This work proposes the biologically plausible spatial adjustment (BSPA) that only calculates the gradient of the neuron at the moment of spiking to the membrane potential. It also proposes the biologically plausible temporal adjustment (BSTA). Spikes that occur affect the spikes that will occur in the future, and the backpropagation algorithm does not consider the influence between the spikes \cite{shen_backpropagation_2022}. This work improves this process by adding a residual connection between the spikes during the backward path.

\subsection{ANN-to-SNN Network} \label{sec:model-conversion}

Since deep neural networks have had major success, and algorithms and tools for their training and evaluation are developed and established in recent years, some researchers have looked for a way to bring this knowledge to spiking neural networks. More specifically, their aim has been to convert trained DNN models to SNNs for inference. Initially, attempts at this conversion resulted in a big accuracy loss. However, this accuracy gap has been filled since by developing new methods. The Diehl et al. paper \cite{diehl_fast-classifying_2015} proposes a conversion scheme. To have minimal conversion loss, the architecture of the DNN network must have some conditions. First, the closest mechanism to the integrate-and-fire activation function in SNNs is rectified linear unit (ReLU) function. Using a ReLU activation function in the DNN structure helps avoid challenges related to negative values and biases \cite{diehl_fast-classifying_2015}. Dropout \cite{srivastava_dropout_2014} layers can be used to reduce problems related to overfitting and the conversion accuracy loss following it.

The paper introduces a weight normalization method that can be used to have near-lossless and faster convergence. Two methods are proposed in this work for weight normalization. The first method considers all positive activations that could occur as input to a layer, and all weights are rescaled by the maximum possible positive input. It is trivial to see this method requires a considerable amount of time to complete. The second method records all the activation values in the ReLU layers for the training set, and after the training is completed, the maximum value is extracted. This method can be faster but makes the decision solely based on the training set. The network used in this work is made of two convolutional layers, each followed by a pooling operation and a fully connected layer at the end. This work achieves an excellent testing accuracy of 99.14\%, but the conversion scheme eliminates biological plausibility.

Another work that improved the ANN-to-SNN process is the work of Ding et al. \cite{ding_optimal_2021}. The problem with previous methods was that the weight normalization process relied on empirical methods and could only happen after the training process. This work proposes a layer called Rate Norm Layer (RNL) to replace the ReLU layer in the ANN. This layer uses a clip function with a trainable upper bound to output the simulated firing rate \cite{ding_optimal_2021}. These trainable parameters can not be learned through the original loss function. Thus, a new term is added to the original loss function of the network.

\section{Conclusion} \label{sec:conclusion}

Deep neural networks have advanced greatly in recent years. They have achieved amazing results and have become state-of-the-art models in some fields of machine learning. However, these networks have shortcomings, such as high energy usage and requiring a large amount of data to learn. The human brain does not have these shortcomings and is able to solve complicated tasks. Although deep neural networks try to model the brain, there are differences between the two. Spiking neural networks are the next generation of neural networks that model the brain better compared to deep neural networks. These networks process spike trains, use more plausible learning rules and model biological neurons better. More attention has been paid to these networks, and researchers have developed several models and structures for these networks. The proposed structures vary in their building blocks, learning rules, and structures.

In this paper, we explained the neural models, coding schemes, learning rules, and simulation tools of spiking neural networks. We listed recent and successful proposed structures for these networks and explored the better-performing ones further.

Having a network that processes binary input has some advantages. These networks have the potential to be highly energy efficient, and this efficiency makes these networks suitable to be implemented on specialized hardware. Furthermore, replacing a good percentage of floating-point operations with integer ones can yield faster runtimes.

Despite the recent progress and developments in the field of spiking neural networks, these networks are not yet comparable to deep neural networks in terms of accuracy. These networks have been able to solve a subset of machine learning problems well, but this success must be spread to other areas and harder problems. This can happen by studying the brain and finding what makes it so great, introducing more complicated dynamics to the networks, and finding structures that perform better.


\begin{thebibliography}{10}

\bibitem{mcculloch_logical_1943}
Warren~S. McCulloch and Walter Pitts.
\newblock A logical calculus of the ideas immanent in nervous activity.
\newblock {\em Bulletin of Mathematical Biophysics}, 5(4):115--133, December
  1943.

\bibitem{rumelhart_learning_1986}
David~E. Rumelhart, Geoffrey~E. Hinton, and Ronald~J. Williams.
\newblock Learning representations by back-propagating errors.
\newblock {\em Nature}, 323(6088):533--536, October 1986.

\bibitem{nair_rectified_2010}
Vinod Nair and Geoffrey~E. Hinton.
\newblock Rectified linear units improve restricted boltzmann machines.
\newblock In {\em Proceedings of the 27th {International} {Conference} on
  {International} {Conference} on {Machine} {Learning}}, {ICML}'10, pages
  807--814, Madison, WI, USA, June 2010. Omnipress.

\bibitem{collobert_natural_2011}
Ronan Collobert, Jason Weston, Léon Bottou, Michael Karlen, Koray Kavukcuoglu,
  and Pavel Kuksa.
\newblock Natural {Language} {Processing} ({Almost}) from {Scratch}.
\newblock {\em Journal of Machine Learning Research}, 12(76):2493--2537, 2011.

\bibitem{graves_framewise_2005}
Alex Graves and Jürgen Schmidhuber.
\newblock Framewise phoneme classification with bidirectional {LSTM} and other
  neural network architectures.
\newblock {\em Neural Networks}, 18(5):602--610, July 2005.

\bibitem{krizhevsky_imagenet_2017}
Alex Krizhevsky, Ilya Sutskever, and Geoffrey~E. Hinton.
\newblock {ImageNet} classification with deep convolutional neural networks.
\newblock {\em Commun. ACM}, 60(6):84--90, May 2017.

\bibitem{sun_revisiting_2017}
Chen Sun, Abhinav Shrivastava, Saurabh Singh, and Abhinav Gupta.
\newblock Revisiting {Unreasonable} {Effectiveness} of {Data} in {Deep}
  {Learning} {Era}.
\newblock In {\em 2017 {IEEE} {International} {Conference} on {Computer}
  {Vision} ({ICCV})}, pages 843--852, October 2017.

\bibitem{li_evaluating_2016}
D.~Li, X.~Chen, M.~Becchi, and Z.~Zong.
\newblock Evaluating the {Energy} {Efficiency} of {Deep} {Convolutional}
  {Neural} {Networks} on {CPUs} and {GPUs}.
\newblock In {\em 2016 {IEEE} {International} {Conferences} on {Big} {Data} and
  {Cloud} {Computing} ({BDCloud}), {Social} {Computing} and {Networking}
  ({SocialCom}), {Sustainable} {Computing} and {Communications} ({SustainCom})
  ({BDCloud}-{SocialCom}-{SustainCom})}, pages 477--484, October 2016.

\bibitem{alcorn_strike_2019}
Michael~A. Alcorn, Qi~Li, Zhitao Gong, Chengfei Wang, Long Mai, Wei-Shinn Ku,
  and Anh Nguyen.
\newblock Strike ({With}) a {Pose}: {Neural} {Networks} {Are} {Easily} {Fooled}
  by {Strange} {Poses} of {Familiar} {Objects}.
\newblock In {\em 2019 {IEEE}/{CVF} {Conference} on {Computer} {Vision} and
  {Pattern} {Recognition} ({CVPR})}, pages 4840--4849, June 2019.

\bibitem{dehaqani_temporal_2016}
Mohammad-Reza~A. Dehaqani, Abdol-Hossein Vahabie, Roozbeh Kiani, Majid~Nili
  Ahmadabadi, Babak~Nadjar Araabi, and Hossein Esteky.
\newblock Temporal dynamics of visual category representation in the macaque
  inferior temporal cortex.
\newblock {\em Journal of Neurophysiology}, 116(2):587--601, August 2016.
\newblock Publisher: American Physiological Society.

\bibitem{dehaqani_selective_2018}
Mohammad-Reza~A Dehaqani, Abdol-Hossein Vahabie, Mohammadbagher Parsa, Behrad
  Noudoost, and Alireza Soltani.
\newblock Selective {Changes} in {Noise} {Correlations} {Contribute} to an
  {Enhanced} {Representation} of {Saccadic} {Targets} in {Prefrontal}
  {Neuronal} {Ensembles}.
\newblock {\em Cerebral Cortex}, 28(8):3046--3063, August 2018.

\bibitem{beer_why_2020}
Michael Beer, Julio Urenda, Olga Kosheleva, and Vladik Kreinovich.
\newblock Why {Spiking} {Neural} {Networks} {Are} {Efficient}: {A} {Theorem}.
\newblock In Marie-Jeanne Lesot, Susana Vieira, Marek~Z. Reformat, João~Paulo
  Carvalho, Anna Wilbik, Bernadette Bouchon-Meunier, and Ronald~R. Yager,
  editors, {\em Information {Processing} and {Management} of {Uncertainty} in
  {Knowledge}-{Based} {Systems}}, Communications in {Computer} and
  {Information} {Science}, pages 59--69, Cham, 2020. Springer International
  Publishing.

\bibitem{abbott_lapicques_1999}
L.~F. Abbott.
\newblock Lapicque's introduction of the integrate-and-fire model neuron
  (1907).
\newblock {\em Brain Research Bulletin}, 50(5-6):303--304, December 1999.

\bibitem{jolivet_spike_2003}
Renaud Jolivet, Timothy J., and Wulfram Gerstner.
\newblock The {Spike} {Response} {Model}: {A} {Framework} to {Predict}
  {Neuronal} {Spike} {Trains}.
\newblock In Okyay Kaynak, Ethem Alpaydin, Erkki Oja, and Lei Xu, editors, {\em
  Artificial {Neural} {Networks} and {Neural} {Information} {Processing} —
  {ICANN}/{ICONIP} 2003}, Lecture {Notes} in {Computer} {Science}, pages
  846--853, Berlin, Heidelberg, 2003. Springer.

\bibitem{izhikevich_simple_2003}
E.M. Izhikevich.
\newblock Simple model of spiking neurons.
\newblock {\em IEEE Transactions on Neural Networks}, 14(6):1569--1572,
  November 2003.
\newblock Conference Name: IEEE Transactions on Neural Networks.

\bibitem{hodgkin_quantitative_1952}
A.~L. Hodgkin and A.~F. Huxley.
\newblock A quantitative description of membrane current and its application to
  conduction and excitation in nerve.
\newblock {\em The Journal of Physiology}, 117(4):500--544, August 1952.

\bibitem{kostakos_human_2017}
Vassilis Kostakos, Jakob Rogstadius, Denzil Ferreira, Simo Hosio, and Jorge
  Goncalves.
\newblock Human {Sensors}.
\newblock In Vittorio Loreto, Muki Haklay, Andreas Hotho, Vito~D.P. Servedio,
  Gerd Stumme, Jan Theunis, and Francesca Tria, editors, {\em Participatory
  {Sensing}, {Opinions} and {Collective} {Awareness}}, Understanding {Complex}
  {Systems}, pages 69--92. Springer International Publishing, Cham, 2017.

\bibitem{adrian_impulses_1926}
E.~D. Adrian and Yngve Zotterman.
\newblock The impulses produced by sensory nerve endings.
\newblock {\em The Journal of Physiology}, 61(4):465--483, August 1926.

\bibitem{wiener_decoding_2003}
Matthew~C. Wiener and Barry~J. Richmond.
\newblock Decoding spike trains instant by instant using order statistics and
  the mixture-of-{Poissons} model.
\newblock {\em The Journal of Neuroscience: The Official Journal of the Society
  for Neuroscience}, 23(6):2394--2406, March 2003.

\bibitem{buzsaki_rhythms_2006}
György Buzsáki.
\newblock {\em Rhythms of the {Brain}}.
\newblock Oxford University Press, New York, 2006.

\bibitem{kheradpisheh_stdp-based_2018}
Saeed~Reza Kheradpisheh, Mohammad Ganjtabesh, Simon~J. Thorpe, and Timothée
  Masquelier.
\newblock {STDP}-based spiking deep convolutional neural networks for object
  recognition.
\newblock {\em Neural Networks}, 99:56--67, March 2018.

\bibitem{mozafari_bio-inspired_2019}
Milad Mozafari, Mohammad Ganjtabesh, Abbas Nowzari-Dalini, Simon~J. Thorpe, and
  Timothée Masquelier.
\newblock Bio-inspired digit recognition using reward-modulated
  spike-timing-dependent plasticity in deep convolutional networks.
\newblock {\em Pattern Recognition}, 94:87--95, October 2019.

\bibitem{johansson_first_2004}
Roland~S. Johansson and Ingvars Birznieks.
\newblock First spikes in ensembles of human tactile afferents code complex
  spatial fingertip events.
\newblock {\em Nature Neuroscience}, 7(2):170--177, February 2004.

\bibitem{thorpe_rank_1998}
Simon Thorpe and Jacques Gautrais.
\newblock Rank {Order} {Coding}.
\newblock pages 113--118, December 1998.

\bibitem{gollisch_rapid_2008}
Tim Gollisch and Markus Meister.
\newblock Rapid {Neural} {Coding} in the {Retina} with {Relative} {Spike}
  {Latencies}.
\newblock {\em Science}, 319(5866):1108--1111, February 2008.

\bibitem{gautrais_rate_1998}
J.~Gautrais and S.~Thorpe.
\newblock Rate coding versus temporal order coding: a theoretical approach.
\newblock {\em Bio Systems}, 48(1-3):57--65, 1998.

\bibitem{okeefe_phase_1993}
J.~O'Keefe and M.~L. Recce.
\newblock Phase relationship between hippocampal place units and the {EEG}
  theta rhythm.
\newblock {\em Hippocampus}, 3(3):317--330, July 1993.

\bibitem{laurent_dynamical_1996}
Gilles Laurent.
\newblock Dynamical representation of odors by oscillating and evolving neural
  assemblies.
\newblock {\em Trends in Neurosciences}, 19(11):489--496, November 1996.

\bibitem{tan_efficientnet_2020}
Mingxing Tan and Quoc~V. Le.
\newblock {EfficientNet}: {Rethinking} {Model} {Scaling} for {Convolutional}
  {Neural} {Networks}, September 2020.

\bibitem{citri_synaptic_2008}
Ami Citri and Robert~C. Malenka.
\newblock Synaptic {Plasticity}: {Multiple} {Forms}, {Functions}, and
  {Mechanisms}.
\newblock {\em Neuropsychopharmacology}, 33(1):18--41, January 2008.
\newblock Number: 1 Publisher: Nature Publishing Group.

\bibitem{markram_regulation_1997}
H.~Markram, J.~Lübke, M.~Frotscher, and B.~Sakmann.
\newblock Regulation of synaptic efficacy by coincidence of postsynaptic {APs}
  and {EPSPs}.
\newblock {\em Science (New York, N.Y.)}, 275(5297):213--215, January 1997.

\bibitem{vreeken_spiking_2003}
Jilles Vreeken.
\newblock Spiking neural networks, an introduction.
\newblock {\em undefined}, 2003.

\bibitem{hebb_organization_1949}
D.~O. Hebb.
\newblock {\em The organization of behavior; a neuropsychological theory}.
\newblock The organization of behavior; a neuropsychological theory. Wiley,
  Oxford, England, 1949.
\newblock Pages: xix, 335.

\bibitem{fremaux_neuromodulated_2016}
Nicolas Frémaux and Wulfram Gerstner.
\newblock Neuromodulated {Spike}-{Timing}-{Dependent} {Plasticity}, and
  {Theory} of {Three}-{Factor} {Learning} {Rules}.
\newblock {\em Frontiers in Neural Circuits}, 9:85, January 2016.

\bibitem{bohte_spikeprop_2000}
Sander Bohte, Joost Kok, and Johannes Poutré.
\newblock {SpikeProp}: backpropagation for networks of spiking neurons.
\newblock volume~48, pages 419--424, January 2000.

\bibitem{diehl_fast-classifying_2015}
Peter~U. Diehl, Daniel Neil, Jonathan Binas, Matthew Cook, Shih-Chii Liu, and
  Michael Pfeiffer.
\newblock Fast-classifying, high-accuracy spiking deep networks through weight
  and threshold balancing.
\newblock In {\em 2015 {International} {Joint} {Conference} on {Neural}
  {Networks} ({IJCNN})}, pages 1--8, July 2015.

\bibitem{paszke_pytorch_2019}
Adam Paszke, Sam Gross, Francisco Massa, Adam Lerer, James Bradbury, Gregory
  Chanan, Trevor Killeen, Zeming Lin, Natalia Gimelshein, Luca Antiga, Alban
  Desmaison, Andreas Kopf, Edward Yang, Zachary DeVito, Martin Raison, Alykhan
  Tejani, Sasank Chilamkurthy, Benoit Steiner, Lu~Fang, Junjie Bai, and Soumith
  Chintala.
\newblock {PyTorch}: {An} {Imperative} {Style}, {High}-{Performance} {Deep}
  {Learning} {Library}.
\newblock In {\em Advances in {Neural} {Information} {Processing} {Systems}},
  volume~32. Curran Associates, Inc., 2019.

\bibitem{developers_tensorflow_2022}
TensorFlow Developers.
\newblock {TensorFlow}, May 2022.

\bibitem{chen_mxnet_2015}
Tianqi Chen, Mu~Li, Yutian Li, Min Lin, Naiyan Wang, Minjie Wang, Tianjun Xiao,
  Bing Xu, Chiyuan Zhang, and Zheng Zhang.
\newblock {MXNet}: {A} {Flexible} and {Efficient} {Machine} {Learning}
  {Library} for {Heterogeneous} {Distributed} {Systems}.
\newblock {\em arXiv:1512.01274 [cs]}, December 2015.

\bibitem{hazan_bindsnet_2018}
Hananel Hazan, Daniel~J. Saunders, Hassaan Khan, Devdhar Patel, Darpan~T.
  Sanghavi, Hava~T. Siegelmann, and Robert Kozma.
\newblock {BindsNET}: {A} {Machine} {Learning}-{Oriented} {Spiking} {Neural}
  {Networks} {Library} in {Python}.
\newblock {\em Front. Neuroinform.}, 12, 2018.

\bibitem{mozafari_spyketorch_2019}
Milad Mozafari, Mohammad Ganjtabesh, Abbas Nowzari-Dalini, and Timothée
  Masquelier.
\newblock {SpykeTorch}: {Efficient} {Simulation} of {Convolutional} {Spiking}
  {Neural} {Networks} {With} at {Most} {One} {Spike} per {Neuron}.
\newblock {\em Front. Neurosci.}, 13, 2019.

\bibitem{pehle_norse_2021}
Christian-Gernot Pehle and Jens~Egholm Pedersen.
\newblock Norse - {A} deep learning library for spiking neural networks,
  January 2021.

\bibitem{stimberg_brian_2019}
Marcel Stimberg, Romain Brette, and Dan~FM Goodman.
\newblock Brian 2, an intuitive and efficient neural simulator.
\newblock {\em eLife}, 8:e47314, August 2019.

\bibitem{gewaltig_nest_2007}
Marc-Oliver Gewaltig and Markus Diesmann.
\newblock {NEST} ({NEural} {Simulation} {Tool}).
\newblock {\em Scholarpedia}, 2(4):1430, April 2007.

\bibitem{shirsavar_faster_2022}
Shahriar~Rezghi Shirsavar and Mohammad-Reza~A. Dehaqani.
\newblock A {Faster} {Approach} to {Spiking} {Deep} {Convolutional} {Neural}
  {Networks}, October 2022.

\bibitem{drix_sparse_2020}
Damien Drix, Verena~V. Hafner, and Michael Schmuker.
\newblock Sparse coding with a somato-dendritic rule.
\newblock {\em Neural Networks}, 131:37--49, November 2020.

\bibitem{xu_improving_2017}
Zihan Xu, Steven Skorheim, Ming Tu, Visar Berisha, Shimeng Yu, Jae-sun Seo,
  Maxim Bazhenov, and Yu~Cao.
\newblock Improving efficiency in sparse learning with the feedforward
  inhibitory motif.
\newblock {\em Neurocomputing}, 267:141--151, December 2017.

\bibitem{saunders_locally_2019}
Daniel~J. Saunders, Devdhar Patel, Hananel Hazan, Hava~T. Siegelmann, and
  Robert Kozma.
\newblock Locally connected spiking neural networks for unsupervised feature
  learning.
\newblock {\em Neural Networks}, 119:332--340, November 2019.

\bibitem{beyeler_categorization_2013}
Michael Beyeler, Nikil~D. Dutt, and Jeffrey~L. Krichmar.
\newblock Categorization and decision-making in a neurobiologically plausible
  spiking network using a {STDP}-like learning rule.
\newblock {\em Neural Networks}, 48:109--124, December 2013.

\bibitem{xu_deep_2020}
Qi~Xu, Jianxin Peng, Jiangrong Shen, Huajin Tang, and Gang Pan.
\newblock Deep {CovDenseSNN}: {A} hierarchical event-driven dynamic framework
  with spiking neurons in noisy environment.
\newblock {\em Neural Networks}, 121:512--519, January 2020.

\bibitem{wang_compsnn_2021}
Tengxiao Wang, Cong Shi, Xichuan Zhou, Yingcheng Lin, Junxian He, Ping Gan,
  Ping Li, Ying Wang, Liyuan Liu, Nanjian Wu, and Gang Luo.
\newblock {CompSNN}: {A} lightweight spiking neural network based on
  spatiotemporally compressive spike features.
\newblock {\em Neurocomputing}, 425:96--106, February 2021.

\bibitem{ding_optimal_2021}
Jianhao Ding, Zhaofei Yu, Yonghong Tian, and Tiejun Huang.
\newblock Optimal {ANN}-{SNN} {Conversion} for {Fast} and {Accurate}
  {Inference} in {Deep} {Spiking} {Neural} {Networks}, May 2021.
\newblock arXiv:2105.11654 [cs].

\vfill
\pagebreak

\bibitem{rueckauer_conversion_2017}
Bodo Rueckauer, Iulia-Alexandra Lungu, Yuhuang Hu, Michael Pfeiffer, and
  Shih-Chii Liu.
\newblock Conversion of {Continuous}-{Valued} {Deep} {Networks} to {Efficient}
  {Event}-{Driven} {Networks} for {Image} {Classification}.
\newblock {\em Frontiers in Neuroscience}, 11, 2017.

\bibitem{zhang_fast_2019}
Anguo Zhang, Hongjun Zhou, Xiumin Li, and Wei Zhu.
\newblock Fast and robust learning in {Spiking} {Feed}-forward {Neural}
  {Networks} based on {Intrinsic} {Plasticity} mechanism.
\newblock {\em Neurocomputing}, 365:102--112, November 2019.

\bibitem{shen_backpropagation_2022}
Guobin Shen, Dongcheng Zhao, and Yi~Zeng.
\newblock Backpropagation with biologically plausible spatiotemporal adjustment
  for training deep spiking neural networks.
\newblock {\em Patterns}, 3(6):100522, June 2022.

\bibitem{lee_enabling_2020}
Chankyu Lee, Syed~Shakib Sarwar, Priyadarshini Panda, Gopalakrishnan
  Srinivasan, and Kaushik Roy.
\newblock Enabling {Spike}-{Based} {Backpropagation} for {Training} {Deep}
  {Neural} {Network} {Architectures}.
\newblock {\em Frontiers in Neuroscience}, 14, 2020.

\bibitem{zhao_backeisnn_2021}
Dongcheng Zhao, Yi~Zeng, and Yang Li.
\newblock {BackEISNN}: {A} {Deep} {Spiking} {Neural} {Network} with {Adaptive}
  {Self}-{Feedback} and {Balanced} {Excitatory}-{Inhibitory} {Neurons}, May
  2021.
\newblock arXiv:2105.13004 [cs].

\bibitem{zhang_spike-train_2019}
Wenrui Zhang and Peng Li.
\newblock Spike-{Train} {Level} {Backpropagation} for {Training} {Deep}
  {Recurrent} {Spiking} {Neural} {Networks}, November 2019.
\newblock arXiv:1908.06378 [cs].

\bibitem{zhang_temporal_2020}
Wenrui Zhang and Peng Li.
\newblock Temporal {Spike} {Sequence} {Learning} via {Backpropagation} for
  {Deep} {Spiking} {Neural} {Networks}.
\newblock In {\em Advances in {Neural} {Information} {Processing} {Systems}},
  volume~33, pages 12022--12033. Curran Associates, Inc., 2020.

\bibitem{wu_spatio-temporal_2018}
Yujie Wu, Lei Deng, Guoqi Li, Jun Zhu, and Luping Shi.
\newblock Spatio-{Temporal} {Backpropagation} for {Training}
  {High}-{Performance} {Spiking} {Neural} {Networks}.
\newblock {\em Frontiers in Neuroscience}, 12, 2018.

\bibitem{cheng_lisnn_2020}
Xiang Cheng, Yunzhe Hao, Jiaming Xu, and Bo~Xu.
\newblock {LISNN}: {Improving} {Spiking} {Neural} {Networks} with {Lateral}
  {Interactions} for {Robust} {Object} {Recognition}.
\newblock volume~2, pages 1519--1525, July 2020.
\newblock ISSN: 1045-0823.

\bibitem{lee_training_2016}
Jun~Haeng Lee, Tobi Delbruck, and Michael Pfeiffer.
\newblock Training {Deep} {Spiking} {Neural} {Networks} {Using}
  {Backpropagation}.
\newblock {\em Front. Neurosci.}, 10, 2016.

\bibitem{kheradpisheh_temporal_2020}
Saeed~Reza Kheradpisheh and Timothée Masquelier.
\newblock Temporal {Backpropagation} for {Spiking} {Neural} {Networks} with
  {One} {Spike} per {Neuron}.
\newblock {\em Int. J. Neur. Syst.}, 30(06):2050027, June 2020.

\bibitem{kheradpisheh_bs4nn_2022}
Saeed~Reza Kheradpisheh, Maryam Mirsadeghi, and Timothée Masquelier.
\newblock {BS4NN}: {Binarized} {Spiking} {Neural} {Networks} with {Temporal}
  {Coding} and {Learning}.
\newblock {\em Neural Process Lett}, 54(2):1255--1273, April 2022.

\bibitem{kingma_adam_2017}
Diederik~P. Kingma and Jimmy Ba.
\newblock Adam: {A} {Method} for {Stochastic} {Optimization}, January 2017.

\bibitem{srivastava_dropout_2014}
Nitish Srivastava, Geoffrey Hinton, Alex Krizhevsky, Ilya Sutskever, and Ruslan
  Salakhutdinov.
\newblock Dropout: {A} {Simple} {Way} to {Prevent} {Neural} {Networks} from
  {Overfitting}.
\newblock {\em Journal of Machine Learning Research}, 15(56):1929--1958, 2014.

\end{thebibliography}
\end{document}